%% file: main.tex
\documentclass[letterpaper]{article} 
\usepackage{aaai2026}  
\usepackage{times}  
\usepackage{helvet}  
\usepackage{courier}  
\usepackage[hyphens]{url}  
\usepackage{graphicx} 
\usepackage{natbib}  
\usepackage{caption} 
\usepackage{algorithm}

\usepackage{newfloat}
\usepackage{listings}
\DeclareCaptionStyle{ruled}{labelfont=normalfont,labelsep=colon,strut=off} 
\floatstyle{ruled}
\newfloat{listing}{tb}{lst}{}
\floatname{listing}{Listing}
\usepackage{booktabs}
\usepackage[table]{xcolor}
\usepackage[olditem]{paralist}
\usepackage{stmaryrd}
\usepackage[noend]{algpseudocode}
\usepackage{multirow}
\usepackage{xspace}
\usepackage{csquotes}
\usepackage{array}
\usepackage{rotating}
\usepackage{tikz}
\usepackage{amsmath, amssymb, amsthm}
\usepackage{xparse}

\newtheorem{definition}{Definition}
\theoremstyle{definition}

\nocopyright 

\title{
   LLM-Only PDDL Domain Repair with Open-Weight Models
}
\author{
    Nader Karimi Bavandpour, Pascal Bercher
}
\affiliations{
    School of Computing, The Australian National University, Canberra, Australia\\
    \{nader.karimiBavandpour, pascal.bercher\}@anu.edu.au

}

\usepackage{bibentry}

\input{macros.tex} 

\begin{document}

\maketitle
\begin{abstract}
AI planning is concerned with finding a sequence of actions that achieves a specified goal. It relies on explicit models of the world, commonly represented in the Planning Domain Definition Language (PDDL). An active line of research investigates how errors in such models can be detected and repaired. For example, users may provide positive test plans that are solutions, and negative test plans that fail during execution. Automated repair methods then modify the PDDL model to satisfy these constraints. In this paper, we evaluate the ability of recent open-weight large language models to perform this repair task using an LLM-only approach. Our experiments show that the symbolic baseline achieves an $F_1$ score of $.49$, while the best-performing LLM reaches $.87$ with high reasoning effort, an absolute improvement of $.38$. However, that setting has a mean test pass rate of only $.82$, falling to $.06$ on the Thoughtful domain; even the best setting that includes the test traces reaches only $.92$. Thus, current open-weight models cannot guarantee satisfaction of the test constraints required for reliable automated model repair.

\end{abstract}



\section{Introduction}
\input{sec-intro}

\section{Planning Formalism}

\input{sec-preliminary}

\section{The Repair Problem}
\input{sec-repair-problems}

\section{Solving the Repair Problem}

\input{sec-solving}

\section{Experiments}

\input{sec-exp}

\section{Conclusion \& Future Work}

\input{sec-conc}

\section{Acknowledgments}
Pascal Bercher is the recipient of an Australian Research Council (ARC) Discovery Early Career Researcher Award (DECRA), project number DE240101245, funded by the Australian Government.

\bibliography{aaai2026}

\onecolumn
\section{Appendix: Per-Domain Results}
Tables~\ref{tab:new-all-twin-models-notrace},
\ref{tab:new-all-twin-models-trace},
\ref{tab:new-all-twin-models-notrace-high}, and
\ref{tab:new-all-twin-models-trace-high} provide the per-domain results for open-weight models: \\
DSV4-F (DeepSeek V4 Flash), DSV4-P (DeepSeek V4 Pro), GLM-5.2 (GLM 5.2), OSS20B (gpt-oss-20B), OSS120B (gpt-oss-120B), Minst-14B (Ministral 3 14B), MistS-4 (Mistral Small 4), MistM-3.5 (Mistral Medium 3.5), MistL-3 (Mistral Large 3), Nem3-S (NVIDIA Nemotron 3 Super), Nem3-U (NVIDIA Nemotron 3 Ultra), Phi4 (Phi-4), Qwen3-30B (Qwen3-30B-A3B), Qwen3-32B. Best values per row are bold; the final two rows report the mean and standard deviation across domains.
\input{Tabs/notrace.tex}
\input{Tabs/trace.tex}

\input{Tabs/notrace_reasoning.tex}
\input{Tabs/trace_reasoning.tex}

\end{document}

%% file: macros.tex
\newdimen\indexdigits
\setbox0\hbox{999}
\indexdigits\wd0
\newcommand\padthreedigits[1]{\hbox to \indexdigits{\hfill#1}}

\newcommand{\domain}{\ensuremath{\mathcal{D}}}
\newcommand{\task}{\ensuremath{\mathcal{T}}}

\newcommand{\objs}{\ensuremath{\mathcal{O}}}
\newcommand{\types}{\ensuremath{\Theta}}
\newcommand{\vars}{\ensuremath{\mathcal{V}}}

\newcommand{\preds}{\ensuremath{\mathcal{P}}}

\newcommand{\facts}{\ensuremath{\mathcal{F}}}

\newcommand{\acts}{\ensuremath{\mathcal{A}}}
\newcommand{\act}{\ensuremath{\mathbf{a}}}
\newcommand{\eff}{\ensuremath{\mathit{eff}}}

\newcommand{\precond}{\ensuremath{\mathit{prec}}}

\newcommand{\preCondPos}[2][]{\ensuremath{\precond^{+}_{#1}(#2)}}
\newcommand{\preCondNeg}[2][]{\ensuremath{\precond^{-}_{#1}(#2)}}
\newcommand{\posEff}[2][]{\ensuremath{\eff^{+}_{#1}(#2)}}
\newcommand{\negEff}[2][]{\ensuremath{\eff^{-}_{#1}(#2)}}

\newcommand{\planGround}{\ensuremath{\gamma}}

\NewDocumentCommand{\repair}{ O{r} O{\mathbf{a}} O{\mathbf{p}} O{c} O{op} }{%
  \ensuremath{#1\llbracket \mathbf{#2}, \mathbf{#3}, #4, #5 \rrbracket}%
}

\newcommand{\subsPartial}{\ensuremath{\varrho}}

\newcommand{\notrace}{\textsc{NoTrace}}
\newcommand{\withtrace}{\textsc{WithTrace}}

%% file: sec-intro.tex
Explainability is a central requirement for AI systems that interact with or support humans in decision making. In AI planning, this requirement is naturally addressed by the explicit representation of actions, states, and goals: planners generate solutions by reasoning over structured models of the world. Compared to black-box machine learning techniques, this explicit reasoning process makes planning inherently transparent and interpretable. However, one of the main challenges to deploying planning in practice lies in constructing the planning models themselves \cite{Muise2025LLMSurvey}.

The recent success of large language models (LLMs) has motivated extensive research into their application to AI planning tasks \cite{Guan2023Lever,Oswald2024Generator,Huang2025Dark,Katz2025,Huang2025Chasing}.
A recent survey \cite{Muise2025LLMSurvey} highlights the potential of LLMs to support the construction and refinement of planning models. While verifiable planning modules remain the backbone of reliability, robustness, and explainability, LLMs can act as assistants to reduce the manual burden of defining domain models. We believe that one promising avenue is \emph{domain repair}, where the goal is to identify a set of modifications to a domain such that the positive traces become executable and the negative traces become non-executable.

Repairs can themselves be understood as explanations. Following \citeauthor{Miller2019ExAI}'s (\citeyear{Miller2019ExAI}) account of contrastive explanations in the social sciences, a repair answers the question of why a given trace fails in the current domain and provides a counterfactual justification of how the domain could have behaved differently. Each repair is thus not only a technical fix, but also a form of interpretable feedback to the human domain modeler.

Several symbolic approaches enforce executability constraints by automatically modifying PDDL models to satisfy them \cite{Aineto2018FAMA,Gragera2025ActionObservations,arthur2026,Bercher2025ModelRepair}, while other work investigates the computational complexity of such repairs \cite{Lin2021FixHTNModel}.
We build on the setting introduced by \citet{Lin2023RepairingClassicalModels} and \citet{Lin2025Blacklist}, in which planning domains are repaired using positive, or whitelist, traces that must be valid plans and negative, or blacklist, traces that must be rendered inapplicable.
They provide a symbolic algorithm that can find cardinality-minimal solutions that satisfy the blacklist and whitelist constraints.
The algorithm was evaluated using precision and recall by comparing the predicted repairs with the ground-truth corrections for PDDL models that had been corrupted through the random addition or removal of action preconditions and effects.

In this paper, we investigate the ability of recent open-weight large language models of varying sizes to solve the PDDL domain-repair problem and compare their performance with the symbolic optimizer described above.
We evaluate all models on the benchmark suite introduced by \citet{Lin2025Blacklist}, enabling a direct comparison with the symbolic approach.
Our experimental setting follows the LLM-only repair formulation introduced under the same name in our earlier work \cite{bavandpour2025finding}, but extends that study by evaluating a broader range of more recent open-weight models, allowing us to assess the current state of the art in LLM-based PDDL model repair.
In addition to the domain-only \notrace{} prompt used previously, we investigate whether the \withtrace{} prompt, which also provides the LLM with the whitelist and blacklist traces, yields further performance improvements.

%% file: sec-preliminary.tex
Since our focus is on repairing lifted PDDL domains, we introduce the lifted planning formalism. A lifted planning problem is defined as a tuple
$\Pi = (\preds{}, \acts{}, \alpha, \objs{}, s^{I}, s^g)$,
where the domain is $\domain{} = (\preds{}, \acts{}, \alpha)$ and the task is $\task{} = (\objs{}, s^{I}, s^g)$.

\paragraph{Objects, Types, and Variables.}
Let $\objs{}$ be the set of objects in the planning task. 
We consider a set of variables $\vars$, each acting as a placeholder for an object.
The type of a variable $v \in \vars$ is written as $v|t$, where $t \in \types{}$. 
Every type $t \in \types{}$ is associated with the set of objects $\objs{}\llbracket t \rrbracket \subseteq \objs{}$. 
We say that $t \in \types{}$ is a \emph{subtype} of $t' \in \types{}$ iff $\objs{}\llbracket t \rrbracket \subseteq \objs{}\llbracket t' \rrbracket$.

\paragraph{Predicates.}
A predicate $\mathbf{p} = P(v_1|t_1, \ldots, v_k|t_k)$ is defined by a unique name $P$ and a tuple of $k \in \mathbb{N}_0$ typed variables, written as 
$\operatorname{par}(\mathbf{p}) := (v_1|t_1, \ldots, v_k|t_k)$. 
The predicate has zero variables if $k=0$.
We denote by $\preds{}$ the set of all predicates in $\Pi$.

\paragraph{Variable Substitution.}
A variable substitution function $\subsPartial : \vars \rightarrow \objs$ maps each typed variable $v|t$ to an object $\subsPartial(v|t) \in \objs{}\llbracket t \rrbracket$ of the same type $t$.

\paragraph{Facts.}
Given a predicate $\mathbf{p} \in \preds$ and a substitution function $\subsPartial$, a \emph{fact} is obtained by grounding $\mathbf{p}$, that is, by replacing each parameter $(v_1, \ldots, v_k)$ with the corresponding objects given by $\subsPartial$:
$f = \subsPartial(\mathbf{p}) = P(\subsPartial(v_1), \ldots, \subsPartial(v_k))$.
The set of all grounded predicates is denoted by $\facts$, and any set of facts constitutes a \emph{state}.

\paragraph{Action Schemas.}
Let $\acts{}$ denote the set of action schemas.  
An action schema $\mathbf{a} = A(v_1|t_1, \ldots, v_k|t_k)$ is defined by a unique name $A$ and a tuple of $k$ variables, written as $\operatorname{par}(\mathbf{a}) := (v_1|t_1, \ldots, v_k|t_k)$.  
Each schema is associated with a mapping
\[
\alpha(\act{}) = (\preCondPos{\act{}}, \preCondNeg{\act{}}, \posEff{\act{}}, \negEff{\act{}})
\]
representing a tuple of four sets of compatible predicates, as defined below.

\begin{definition}[Compatible Predicates]
	For an action schema $\mathbf{a}$, the set of \emph{compatible predicates} $\mathcal{P}^{\mathbf{a}}$ contains all predicates whose set of parameters is a subset of those of $\mathbf{a}$:
	\[
	\mathcal{P}^{\mathbf{a}} := \{\mathbf{p} \in \preds{} \mid \operatorname{s\text{-}par}(\mathbf{p}) \subseteq \operatorname{s\text{-}par}(\mathbf{a})\},
	\]
	where $\operatorname{s\text{-}par}$ converts the parameters into a set.
\end{definition}

\paragraph{Actions.}
Given an action schema $\mathbf{a}$ and a substitution function $\subsPartial$, the corresponding \emph{action} is obtained by replacing each parameter of $\mathbf{a}$ according to $\subsPartial$, and is denoted $a = \mathbf{a}[\subsPartial]$.
Actions describe transitions in the state space.
An action $a$ is \emph{applicable} in a state $s$ iff $\preCondPos{a} \subseteq s$ and $\preCondNeg{a} \cap s = \emptyset$.
Applying an applicable action $a$ in $s$ produces the successor state
\[
s' = (s \setminus \negEff{a}) \cup \posEff{a},
\]
which we denote by $s \to_{a} s'$.

Throughout this paper, we use boldface (e.g., $\mathbf{p}$, $\mathbf{a}$) for predicates and action schemas, and regular typeface (e.g., $f$, $a$) for facts and actions.

\paragraph{Solutions.}
Let $\planGround = \langle a_{1}, \ldots, a_{k} \rangle$ be an action sequence.
We write $s \rightarrow^{*}_{\planGround} s'$ to denote that $s'$ results from applying $\planGround$ to $s$ via a state trajectory $\langle s_{0}, \ldots, s_{k} \rangle$ where $s_{0} = s$, $s_{k} = s'$, and each action is applicable in its preceding state.
A solution to a planning problem is an action sequence $\planGround = \langle a_1, \ldots, a_k \rangle$ such that $s_I \to^*_\planGround s'$ for some $s'$ with $s^g \subseteq s'$, and each $a_i$ is a grounding of some action schema $\mathbf{a} \in \acts{}$.

%% file: sec-repair-problems.tex
We begin by introducing the notation and syntax used to define possible repair operations for a given planning domain. Next, we describe how a set of such repairs can be applied to produce a modified domain. Based on these concepts, we then formalize the domain repair problem in terms of the defined repair operations and a set of positive and negative plans. Our formulation follows the setting introduced by \citet{Lin2025Blacklist} as closely as possible, while explicitly highlighting the modifications needed to incorporate LLMs for prioritizing semantically meaningful repair sets.

\input{Prompts/main.tex}

In a planning domain $\domain = (\preds, \acts, \alpha)$, an \emph{atomic repair} is a modification denoted by $\repair$. Here, $\mathbf{a} \in \acts$ is an action schema, $\mathbf{p} \in \preds$ is a predicate compatible with $\mathbf{a}$, $c \in \{\mathrm{prec}^+, \mathrm{prec}^-, \mathrm{eff}^+, \mathrm{eff}^-\}$ indicates whether the change concerns a positive or negative precondition or effect, and $op \in \{+, -\}$ specifies whether the component is added or removed. We write $\domain \Rightarrow_r \domain'$ to indicate that applying $r$ to $\domain$ yields $\domain' = (\preds, \acts, \alpha')$, where $\alpha'$ results from applying $r$ to $\alpha$.

A \emph{repair set} $\delta$ for a domain is a finite collection of zero or more atomic repairs.  
We say that $\delta$ is \emph{valid} if and only if it contains no two repairs $r, r' \in \delta$ such that one reverses the effect of the other.  
Specifically, two repairs $r = \repair$ and $r' = \repair[r'][a'][p'][c'][op']$ are considered to undo each other if  
$\mathbf{a} = \mathbf{a}'$, $\mathbf{p} = \mathbf{p}'$, $c = c'$, and $op \neq op'$.

Let $\mathcal{D}$ be a domain and $\delta$ a valid repair set for $\mathcal{D}$.  
Applying the repairs in $\delta$ in any order yields the same modified domain $\mathcal{D}'$.  
We use $\mathcal{D} \Rightarrow_\delta^* \mathcal{D}'$ to indicate that $\mathcal{D}'$ is obtained from $\mathcal{D}$ by applying the valid repair set $\delta$.

\begin{definition}[Domain Repair Problem]
The domain repair problem is defined as a pair $\mathcal{R} = (\domain{}, \mathbb{T})$, where $\domain{}$ denotes a planning domain and $\mathbb{T} = \{\mathbf{T}_1, \dots, \mathbf{T}_n\}$ for some $n \in \mathbb{N}$. Each element $\mathbf{T}_i$ is a triple $(\Pi_i, \mathbb{P}_i, \mathbb{E}_i)$. Here, $\Pi_i=(\domain{},\mathcal{T}_i)$ denotes the planning problem; $\mathbb{P}_i$ is a finite, nonempty set of positive plans $\pi^+_k$ for $\Pi_i$; and $\mathbb{E}_i$ is a finite, nonempty set of pairs $(\pi_k^-, i_k)$ associated with $\Pi_i$. Each $\pi^-_k$ denotes a negative plan (a sequence of actions considered undesirable) for $\Pi_i$, and $i_k$ is at most the length of that plan.
\end{definition}

\begin{definition}[Solution to the Repair Problem]
	A \emph{solution} to $\mathcal{R}$ is a valid repair set $\delta$ that transforms the original domain $\mathcal{D}$ into a modified domain $\mathcal{D}'$ through the sequence of repair operations $\mathcal{D} \Rightarrow_\delta^* \mathcal{D}'$. This repair must satisfy the following conditions: for every index $i$ with $1 \leq i \leq n$, all positive plans $\pi^+ \in \mathbb{P}_i$ must be valid solutions to the updated planning problem $\Pi_i' = (\mathcal{D}', \mathcal{T}_i)$, meaning that they are executable and achieve the goal; and for each pair $(\pi_k^-, i_k) \in \mathbb{E}_i$, $\pi_k^-$ must not be a valid plan for $\Pi_i'$, with the action at position $i_k$ being the first that cannot be applied.
\end{definition}

%% file: Prompts/main.tex
\begin{figure}[!t]
\centering
\fbox{%
\begin{minipage}{\dimexpr\columnwidth-2\fboxsep-2\fboxrule\relax}
\footnotesize
\raggedright
\textbf{Variation 1: \notrace{} (Domain Only).}
\begin{compactenum}
\item Infer the intended semantics of each action in the flawed PDDL domain.
\item Propose plausible single-edit repairs: add or remove a positive or
negative precondition or effect.
\item Return a concise action overview, the recommended repairs grouped by action, and a justification for each repair.
\end{compactenum}

\textbf{Constraints.}
\begin{compactitem}
\item Use semantic cues from action and predicate names.
\item Introduce no new predicates, variables, or constants; added atoms use
arguments available to the action.
\end{compactitem}

\textbf{Input:} a flawed PDDL domain.

\smallskip
\hrule
\smallskip

\textbf{Variation 2: \withtrace{} (Domain and Test Traces).}
\begin{compactitem}
\item Follow all \notrace{} steps and constraints.
\item Also use test traces as behavioral evidence.
\item Positive (whitelist) plans must execute and reach the goal.
\item For a negative (blacklist) plan with failure index $n$, its prefix must
be executable, and action $n$ must be inapplicable.
\end{compactitem}

\textbf{Input:} a flawed PDDL domain and positive/negative test traces.
\end{minipage}%
}
\caption{The \notrace{} and \withtrace{} LLM-only prompt abstractions used in our evaluation.}
\label{fig:prompt-variations}
\end{figure}

%% file: sec-solving.tex
In this section, we briefly review our symbolic baseline and then introduce our LLM-only approach.

\subsection{The Baseline Approach}

The symbolic baseline \cite{Lin2023RepairingClassicalModels, Lin2025Blacklist} uses a sound algorithm based on conditional hitting sets to solve the domain repair problem. The authors report runtime and evaluate precision and recall against known ground-truth repairs. To obtain the ground truth, they perturb IPC domains by randomly adding or removing preconditions and effects, which allows precision and recall to be computed directly.

The algorithm executes all whitelist and blacklist test plans. For each failing test, it identifies the possible repairs that could resolve the failure and encodes them as a conditional hitting-set problem. The solver then selects a minimal repair set, applies it to the domain, and re-evaluates the tests under the modified model. If additional tests fail, the newly identified repair requirements are incorporated and the process continues iteratively until a minimal repair set satisfying all tests is found.

The baseline does not exploit semantic cues encoded by the modeler in predicate, action, or domain names within the PDDL file. Its hitting-set solver optimizes only the size of the repair set; when multiple diagnoses share the same cardinality, it returns an arbitrary one. Moreover, the ground-truth repair need not be cardinality-minimal, so it can be missed under this objective.

\input{Tabs/models.tex}

\subsection{The LLM-Only Approach}
In our previous work \cite{bavandpour2025finding}, we introduced several ideas on how LLMs can be exploited to improve purely symbolic repair approaches, including the LLM-only approach studied here. In this approach, an LLM directly predicts a repair set without symbolic post-processing.
In that study, we evaluated only a single LLM, limiting the scope of the study.
Here, we extend the results for that approach using more recent LLMs.
The present study evaluates whether LLM reasoning can usefully complement symbolic repair.
It also examines whether current LLMs are strong enough for the LLM-only approach to produce repairs of sufficient overall quality.

Specifically, we evaluate the two prompts summarized in Figure~\ref{fig:prompt-variations}. 
The \notrace{} prompt supplies only the corrupted domain and asks the LLM to infer intended preconditions and effects from the action and predicate names.
The \withtrace{} prompt additionally supplies the positive and negative test traces and instructs the LLM to choose repairs that make every positive trace executable and goal-achieving and every negative trace fail at its designated action.

The LLM-only \withtrace{} setting directly addresses the repair problem because the LLM has access to the test constraints, although its output may still be noisy or unsound. In contrast, \notrace{} does not have access to the tests and therefore addresses a relaxed variant of the problem in which no test constraints must be satisfied. Consequently, multiple semantically plausible repairs may be proposed for any action without regard to whether they solve the actual repair instance.

%% file: Tabs/models.tex
\begin{table*}[p]
\centering
\setlength{\tabcolsep}{3pt}
\renewcommand{\arraystretch}{1.2}

\begingroup
\setbox0=\hbox{%
\begin{tabular}{lccccc|cccc|cccc|}
\multicolumn{6}{c}{} &
\multicolumn{4}{c}{Default Effort} &
\multicolumn{4}{c}{High Effort} \\
\cline{7-10}\cline{11-14}
\multicolumn{1}{c}{LLM Model Name} &
\multicolumn{1}{c}{\shortstack{Default\\Effort}} &
\multicolumn{1}{c}{\shortstack{High\\Effort}} &
\multicolumn{1}{c}{Release Date} &
\multicolumn{1}{c}{\shortstack{Active\\Par. (B)}} &
\multicolumn{1}{c}{\shortstack{Total\\Par. (B)}} &
\multicolumn{1}{c}{Pr} &
\multicolumn{1}{c}{Re} &
\multicolumn{1}{c}{$F_1$} &
\multicolumn{1}{c}{TR} &
\multicolumn{1}{c}{Pr} &
\multicolumn{1}{c}{Re} &
\multicolumn{1}{c}{$F_1$} &
\multicolumn{1}{c}{TR} \\
\hline

\rowcolor{gray!20}
Symbolic Baseline
& - & - & - & - & -
& \textbf{\textit{.69}} & .41 & \textbf{\textit{.49}} & $\mathbf{1}^{\ast}$
& - & - & - & - \\

OpenAI GPT-4o
& Unspec. & - & 24-05 & NA & NA
& .48 & .38 & .38 & .27
& - & - & - & - \\

Phi-4
& Unspec. & - & 24-12 & 14 & 14
& .03 & .05 & .04 & .10
& - & - & - & - \\

Qwen3-30B-A3B
& Unspec. & - & 25-04 & 3.3 & 30.5
& .46 & .32 & .34 & .23
& - & - & - & - \\

Qwen3-32B
& Unspec. & Enabled & 25-04 & 32.8 & 32.8
& .52 & .46 & .47 & .50
& .41 & .51 & .40 & .58 \\

OpenAI GPT-OSS-120B
& Medium & High & 25-08 & 5.1 & 117
& .56 & .58 & .53 & .50
& .34 & .44 & .37 & .40 \\

OpenAI GPT-OSS-20B
& Medium & High & 25-08 & 3.6 & 21
& .42 & .51 & .42 & .27
& .47 & .60 & .50 & .50 \\

Ministral 3 14B
& Unspec. & - & 25-12 & 14 & 14
& .23 & .34 & .27 & .12
& - & - & - & - \\

Mistral Large 3
& Unspec. & - & 25-12 & 41 & 675
& .29 & .37 & .31 & .30
& - & - & - & - \\

Mistral Small 4
& Off & High & 26-03 & 6.5 & 119
& .14 & .15 & .13 & .15
& .45 & .51 & .41 & .36 \\

NVIDIA Nemotron 3 Super
& Medium & - & 26-03 & 12 & 120
& .40 & .59 & .43 & .33
& - & - & - & - \\

DeepSeek V4 Flash
& Unspec. & X-High & 26-04 & 13 & 284
& .57 & .73 & .60 & .56
& .67 & .83 & .70 & .72 \\

DeepSeek V4 Pro
& Unspec. & X-High & 26-04 & 49 & 1600
& \textbf{.91} & .76 & .80 & .69
& .61 & \textbf{.89} & .69 & .65 \\

Mistral Medium 3.5
& Unspec. & High & 26-04 & 128 & 128
& .34 & .45 & .35 & .42
& .44 & .72 & .52 & .52 \\

GLM 5.2
& High & X-High & 26-06 & 40 & 753
& .90 & \textbf{.82} & \textbf{.85} & \textbf{.85}
& \textbf{.93} & .83 & \textbf{.87} & \textbf{.82} \\

NVIDIA Nemotron 3 Ultra
& High & - & 26-06 & 55 & 550
& .63 & .76 & .65 & .74
& - & - & - & - \\
\hline
\end{tabular}
}%
\ifdim\wd0>\textwidth
\resizebox{\textwidth}{!}{\usebox0}%
\else
\usebox0
\fi
\endgroup

\par\vspace{6pt}

\begingroup
\setbox0=\hbox{%
\begin{tabular}{lccccc|cccc|cccc|}
\multicolumn{6}{c}{} &
\multicolumn{4}{c}{Default Effort} &
\multicolumn{4}{c}{High Effort} \\
\cline{7-10}\cline{11-14}
\multicolumn{1}{c}{LLM Model Name} &
\multicolumn{1}{c}{\shortstack{Default\\Effort}} &
\multicolumn{1}{c}{\shortstack{High\\Effort}} &
\multicolumn{1}{c}{Release Date} &
\multicolumn{1}{c}{\shortstack{Active\\Par. (B)}} &
\multicolumn{1}{c}{\shortstack{Total\\Par. (B)}} &
\multicolumn{1}{c}{Pr} &
\multicolumn{1}{c}{Re} &
\multicolumn{1}{c}{$F_1$} &
\multicolumn{1}{c}{TR} &
\multicolumn{1}{c}{Pr} &
\multicolumn{1}{c}{Re} &
\multicolumn{1}{c}{$F_1$} &
\multicolumn{1}{c}{TR} \\
\hline

OpenAI GPT-4o
& Unspec. & - & 24-05 & NA & NA
& .43 & .29 & .33 & .31
& - & - & - & - \\

Phi-4
& Unspec. & - & 24-12 & 14 & 14
& 0 & 0 & 0 & .07
& - & - & - & - \\

Qwen3-30B-A3B
& Unspec. & - & 25-04 & 3.3 & 30.5
& .25 & .15 & .18 & .10
& - & - & - & - \\

Qwen3-32B
& Unspec. & Enabled & 25-04 & 32.8 & 32.8
& .26 & .22 & .22 & .23
& .12 & .10 & .11 & .11 \\

OpenAI GPT-OSS-120B
& Medium & High & 25-08 & 5.1 & 117
& .67 & .45 & .49 & .57
& .15 & .16 & .15 & .09 \\

OpenAI GPT-OSS-20B
& Medium & High & 25-08 & 3.6 & 21
& .26 & .24 & .23 & .26
& .38 & .47 & .41 & .55 \\

Ministral 3 14B
& Unspec. & - & 25-12 & 14 & 14
& .09 & .14 & .10 & .17
& - & - & - & - \\

Mistral Large 3
& Unspec. & - & 25-12 & 41 & 675
& .44 & .42 & .35 & .30
& - & - & - & - \\

Mistral Small 4
& Off & High & 26-03 & 6.5 & 119
& .10 & .11 & .09 & .11
& .55 & .49 & .44 & .50 \\

NVIDIA Nemotron 3 Super
& Medium & - & 26-03 & 12 & 120
& .57 & .27 & .34 & .48
& - & - & - & - \\

DeepSeek V4 Flash
& Unspec. & X-High & 26-04 & 13 & 284
& .69 & .64 & .63 & .75
& .81 & .69 & .71 & .81 \\

DeepSeek V4 Pro
& Unspec. & X-High & 26-04 & 49 & 1600
& .85 & \textbf{.71} & .74 & \textbf{.82}
& .85 & \textbf{.83} & \textbf{.82} & \textbf{.92} \\

Mistral Medium 3.5
& Unspec. & High & 26-04 & 128 & 128
& .42 & .44 & .36 & .46
& .74 & .58 & .62 & .75 \\

GLM 5.2
& High & X-High & 26-06 & 40 & 753
& .94 & .70 & \textbf{.78} & .82
& \textbf{.94} & .76 & .82 & .85 \\

NVIDIA Nemotron 3 Ultra
& High & - & 26-06 & 55 & 550
& \textbf{.95} & .66 & .74 & .78
& - & - & - & - \\
\hline
\end{tabular}
}%
\ifdim\wd0>\textwidth
\resizebox{\textwidth}{!}{\usebox0}%
\else
\usebox0
\fi
\endgroup

\caption{\small
The top table shows the \notrace{} and the bottom one shows the \withtrace{} LLM-only results.
Each row reports the metrics averaged over the 12 error-injected IPC domains described in the text. Parameter counts are given in billions. Pr, Re, and $F_1$ denote precision, recall, and their harmonic mean, respectively; TR is the average test pass rate, i.e., the fraction of tests satisfied by the repairs. The grayed-out first row of the top table gives the symbolic-only baseline, whose TR is 1 by construction and is marked $1^{\ast}$. \emph{Default Effort} uses the model's default reasoning configuration, while \emph{High Effort} uses the enhanced configuration specified in that column. \emph{Unspec.} means default reasoning enablement is not advertised. A ``-'' in an effort column means no alternative configuration is supported; elsewhere, it denotes an inapplicable result.
}
\label{tab:new-model-based}
\end{table*}

%% file: sec-exp.tex
We evaluate our LLM-only approach on the benchmark suite introduced by \citet{Lin2025Blacklist} and later used in our previous study \cite{bavandpour2025finding}, enabling comparison under the same precision, recall, and $F_1$ metrics. For a ground-truth repair set $G$ and predicted repair set $P$, precision is $|P \cap G|/|P|$, recall is $|P \cap G|/|G|$, and $F_1$ is their harmonic mean, with all three scores defined as $0$ when $P$ is empty. We additionally report the test pass rate (TR), the fraction of tests satisfied by the predicted repair. TR is particularly important because a solution to the domain repair problem must satisfy every test: high $F_1$ without near-perfect TR does not provide the required correctness and is therefore insufficient for a reliable stand-alone solver. Each domain represents one repair problem, for which we perform one experimental run; all LLM calls were made through OpenRouter\footnote{\url{https://openrouter.ai/}}. If a call fails or its response cannot be parsed as a repair set, we repeat the same request up to three additional times, and an instance that still has no valid response is treated as an empty prediction. These retries recover failed calls rather than constituting additional experimental runs. The AVG and STD rows report, respectively, the unweighted mean and standard deviation over the per-domain results, so the reported averages are macro averages.

We omit the older versions of the Logistics and Woodworking domains (LOGISTICS98 and WOODWORKING08), retaining LOGISTICS00 and WOODWORKING11. We also omit MPRIME because that domain is designed to use misleading names, making it unsuitable for evaluating an LLM's ability to exploit semantic cues.
Because the symbolic method is stochastic, we repeat in Table~\ref{tab:new-model-based} the five-run average published by \citet{Lin2025Blacklist} instead of rerunning it.

Table~\ref{tab:new-model-based} presents our main results for the recent
models.  We repeat the symbolic result in its first row only to provide a
common point of reference.  Compared with the \notrace{} prompt used for the
earlier runs reported in our previous work \cite{bavandpour2025finding}, the \notrace{} prompt in these experiments is shorter: it neither requests an
explicit reasoning trace nor provides a one-shot example.  We made this
change to keep the task and its context simpler for smaller models, whose
performance is a central focus of this evaluation.
The \withtrace{} prompt minimally extends \notrace{} by supplying the whitelist and blacklist test traces and asking the LLM to satisfy all tests.

\input{Tabs/pivot.tex}

Our previous work reports a mean $F_1$ of $.46$ for GPT-4o with the original \notrace{} prompt \cite{bavandpour2025finding}, compared with $.38$ using the shorter prompt in Table~\ref{tab:new-model-based}.
With test traces excluded and default reasoning, GLM 5.2 achieves the highest mean $F_1$ of $.85$.
The release dates in
Table~\ref{tab:new-model-based} also
show a pronounced generational improvement: the best 2025 model reaches
$F_1=.53$, whereas the best 2026 model reaches $F_1=.85$.

Table~\ref{tab:pivot} summarizes the interaction between trace inclusion and
reasoning effort.  Supplying the test traces does not improve the best result
at default effort: GLM 5.2 remains the best model, but its mean $F_1$ falls
from $.85$ to $.78$.  Reasoning jointly over many potentially long plans is
combinatorial and can also consume or exceed the LLM's context window, which
may explain why the additional evidence is not consistently useful.  The
per-domain results support this interpretation: Table~\ref{tab:new-all-twin-models-notrace}
contains a numeric result for every model-domain pair, whereas
Table~\ref{tab:new-all-twin-models-trace} contains both context-limit errors
and invalid or unparsable outputs.

Higher reasoning effort helps several models, but not all of them.  Without
test traces, Table~\ref{tab:new-model-based} shows that GLM 5.2 improves from
$F_1=.85$ to $.87$.
The increase is larger for Mistral Small 4, from $.13$ to $.41$.
Nevertheless, Table~\ref{tab:pivot} shows that providing test traces at high effort yields a
best mean $F_1$ of only $.82$, shared by DeepSeek V4 Pro and GLM 5.2.  This is
below the best default-effort \notrace{} result of $.85$.

Test satisfaction exposes a more consequential limitation.  The strongest
LLM result, GLM 5.2 with high effort and \notrace{}, improves substantially on
the symbolic baseline in $F_1$ ($.87$ versus $.49$), but its mean TR is only
$.82$.  Its TR falls to $.06$ on Thoughtful, meaning that only $6\%$ of that
domain's tests are satisfied.  Thus, high repair-set overlap does not provide
the strict correctness guarantee required by the repair problem.

Reasoning over the test traces does help with their combinatorial constraints
in one important case: for DeepSeek V4 Pro with \withtrace{}, increasing the
reasoning effort raises mean TR from $.82$ to $.92$ while also raising $F_1$
from $.74$ to $.82$.  Nevertheless, $.92$ still falls short of complete test
satisfaction, and TR again drops to $.06$ on Thoughtful.  Recent open-weight
models therefore improve semantic repair quality without guaranteeing that
the predicted repair is a solution.  These findings are supported only by our
error-injected IPC benchmark, which is publicly available online\footnote{\url{https://zenodo.org/records/14533200} \cite{Lin2024ExperimentalResultsRepairingDomains}}.  The models
may therefore have encountered the underlying domains or some of their fixes
during training and could be recalling them rather than deriving every repair
from the supplied instance.  This possible benchmark contamination limits how
strongly the observed $F_1$ and TR gains can be expected to generalize.

We argue for hybrid methods that use an
LLM to rank or filter semantically promising candidates while a symbolic
component provides theoretical test-satisfaction guarantees, as proposed in
our previous work \cite{bavandpour2025finding}; realizing such a method remains
future work.

%% file: Tabs/pivot.tex
\begin{table}[!t]
\centering
\small
\setlength{\tabcolsep}{2.3pt}
\renewcommand{\arraystretch}{1.15}
\begin{tabular}{llccc@{\hspace{6pt}}ccc}
\toprule
&& \multicolumn{3}{c}{Default Effort}
& \multicolumn{3}{c}{High Effort} \\
\cmidrule(lr){3-5}
\cmidrule(lr){6-8}
Prompt
& & $F_1$ & TR & Model
& $F_1$ & TR & Model \\
\midrule

\multirow{2}{*}{\notrace{}}
& AVG & .85 & .85 & GLM-5.2 & \textbf{.87} & .82 & GLM-5.2 \\
& STD & .17 & .28 &         & .12          & .29 &         \\

\cmidrule(lr){2-8}

\multirow{2}{*}{\withtrace{}}
& AVG & .78 & .82 & GLM-5.2 & .82 & \textbf{.92} & DSV4-P \\
& STD & .22 & .29 &         & .17 & .26 &        \\

\bottomrule
\end{tabular}
\caption{Average $F_1$ and TR for each trace-inclusion and reasoning-effort setting, with the corresponding best-performing model.
These metrics are reported as in Table~\ref{tab:new-model-based}; see its caption for their definitions.
DSV4-P: DeepSeek V4 Pro; GLM-5.2: GLM 5.2.
For High Effort with \withtrace{}, GLM-5.2 matches DSV4-P's mean $F_1$, but
DSV4-P is shown because its $F_1$ standard deviation is lower (.17 versus .18).
Per-domain breakdowns are reported in
Tables~\ref{tab:new-all-twin-models-notrace},
\ref{tab:new-all-twin-models-trace},
\ref{tab:new-all-twin-models-notrace-high}, and
\ref{tab:new-all-twin-models-trace-high}.
}
\label{tab:pivot}
\end{table}

%% file: sec-conc.tex
Our results convey two main messages.  First, recent open-weight LLMs can use
semantic cues to identify repairs that agree much more closely with the ground
truth than the symbolic baseline: the best mean $F_1$ is $.87$, compared with
$.49$.  Reasoning effort can strengthen this ability, but its effect is
model-dependent, and supplying test traces does not improve the best $F_1$.
Second, repair-set overlap is not a correctness guarantee.  The setting with
the best $F_1$ has a mean TR of $.82$ and a TR of only $.06$ on Thoughtful;
even the strongest trace-aware setting reaches a mean TR of $.92$ and again
only $.06$ on Thoughtful.  Current LLM-only methods therefore improve
semantic repair quality but cannot reliably solve the repair problem, as
every test must pass.

A promising direction is consequently to combine the complementary strengths
of both approaches: an LLM can express semantic repair preferences, while a
symbolic reasoner preserves theoretical test-satisfaction guarantees
\cite{bavandpour2025finding}.  We will also isolate the effects of prompt
length, explicit reasoning requests, and examples.  Finally, because the
evaluated IPC-derived benchmark is public, possible training-data recall
limits the generality of our results; evaluation on novel, unpublished domains
is needed to establish whether these gains transfer to genuinely unseen repair
problems.

%% file: Tabs/notrace.tex
\begin{table*}[!h]
\centering
\setlength{\tabcolsep}{3pt}
\renewcommand{\arraystretch}{1.2}
\begingroup
\setbox0=\hbox{%
\begin{tabular}{|l|cccc|cccc|cccc|cccc|cccc|cccc|cccc|}
\hline
 & \multicolumn{4}{c|}{OSS20B} & \multicolumn{4}{c|}{OSS120B} & \multicolumn{4}{c|}{Minst-14B} & \multicolumn{4}{c|}{MistL-3} & \multicolumn{4}{c|}{Phi4} & \multicolumn{4}{c|}{Qwen3-30B} & \multicolumn{4}{c|}{Qwen3-32B} \\
 Domain & Pr & Re & $F_1$ & TR & Pr & Re & $F_1$ & TR & Pr & Re & $F_1$ & TR & Pr & Re & $F_1$ & TR & Pr & Re & $F_1$ & TR & Pr & Re & $F_1$ & TR & Pr & Re & $F_1$ & TR \\
\hline
Floortile & .67 & .50 & .57 & .17 & .67 & \textbf{1} & .80 & .83 & .60 & .75 & .67 & .17 & .40 & .50 & .44 & 0 & 0 & 0 & 0 & 0 & .50 & .50 & .50 & 0 & \textbf{1} & .75 & .86 & .17 \\
Freecell & 0 & 0 & 0 & 0 & .25 & .50 & .33 & .21 & .14 & .25 & .18 & .01 & .14 & .25 & .18 & 0 & 0 & 0 & 0 & 0 & .40 & .50 & .44 & .21 & .67 & .50 & .57 & .21 \\
GED & .10 & .10 & .10 & 0 & \textbf{1} & .20 & .33 & .55 & .12 & .20 & .15 & 0 & .18 & .20 & .19 & 0 & 0 & 0 & 0 & 0 & .67 & .20 & .31 & .31 & .20 & .10 & .13 & 0 \\
Hiking & .80 & \textbf{1} & .89 & .22 & .75 & .75 & .75 & .22 & .33 & .50 & .40 & 0 & .75 & .75 & .75 & \textbf{1} & 0 & 0 & 0 & .78 & .67 & .50 & .57 & .22 & .75 & .75 & .75 & \textbf{1} \\
Logistics00 & .67 & \textbf{1} & .80 & \textbf{1} & .67 & \textbf{1} & .80 & \textbf{1} & .17 & .25 & .20 & 0 & .50 & .50 & .50 & 0 & 0 & 0 & 0 & .03 & .67 & .50 & .57 & 0 & \textbf{1} & .75 & .86 & .97 \\
Scanalyzer & .17 & \textbf{1} & .29 & 0 & .67 & \textbf{1} & .80 & \textbf{1} & .25 & .50 & .33 & .40 & 0 & 0 & 0 & 0 & 0 & 0 & 0 & 0 & 0 & 0 & 0 & 0 & .20 & .50 & .29 & .93 \\
Slitherlink & \textbf{1} & .50 & .67 & .33 & 0 & 0 & 0 & 0 & .33 & .50 & .40 & .33 & .33 & .50 & .40 & .67 & .25 & .50 & .33 & .33 & 0 & 0 & 0 & 0 & .50 & .50 & .50 & .33 \\
Sokoban & .25 & .50 & .33 & .40 & \textbf{1} & \textbf{1} & \textbf{1} & \textbf{1} & 0 & 0 & 0 & 0 & 0 & 0 & 0 & 0 & 0 & 0 & 0 & 0 & .33 & .50 & .40 & .40 & 0 & 0 & 0 & 0 \\
Tetris & .67 & .50 & .57 & .50 & .09 & .25 & .13 & .60 & .20 & .50 & .29 & 0 & .19 & \textbf{.75} & .30 & .50 & 0 & 0 & 0 & 0 & .29 & .50 & .36 & .50 & .22 & .50 & .31 & .50 \\
Thoughtful & .09 & .10 & .10 & 0 & .44 & .40 & .42 & \textbf{.06} & .29 & .20 & .24 & 0 & .50 & .40 & .44 & 0 & .12 & .10 & .11 & \textbf{.06} & .33 & .10 & .15 & 0 & .17 & .20 & .18 & 0 \\
Tidybot & .12 & .08 & .10 & .22 & .38 & .25 & .30 & .11 & .14 & .08 & .11 & .44 & .17 & .08 & .11 & .67 & 0 & 0 & 0 & 0 & .67 & .17 & .27 & .44 & .50 & .33 & .40 & \textbf{1} \\
Woodwork11 & .45 & .83 & .59 & .38 & .80 & .67 & .73 & .38 & .22 & .33 & .27 & .12 & .30 & .50 & .37 & .75 & 0 & 0 & 0 & 0 & \textbf{1} & .33 & .50 & .62 & \textbf{1} & .67 & .80 & .88 \\
\hline
AVG & .42 & .51 & .42 & .27 & .56 & .58 & .53 & .50 & .23 & .34 & .27 & .12 & .29 & .37 & .31 & .30 & .03 & .05 & .04 & .10 & .46 & .32 & .34 & .23 & .52 & .46 & .47 & .50 \\
STD & .32 & .37 & .29 & .28 & .32 & .35 & .30 & .37 & .14 & .21 & .16 & .17 & .21 & .25 & .21 & .37 & .07 & .14 & .09 & .22 & .28 & .20 & .19 & .22 & .35 & .25 & .29 & .41 \\
\hline
\end{tabular}
}%
\ifdim\wd0>\textwidth
\resizebox{\textwidth}{!}{\usebox0}%
\else
\usebox0
\fi
\endgroup

\par\vspace{6pt}
\begingroup
\setbox0=\hbox{%
\begin{tabular}{|l|cccc|cccc|cccc|cccc|cccc|cccc|cccc|}
\hline
 & \multicolumn{4}{c|}{DSV4-F} & \multicolumn{4}{c|}{DSV4-P} & \multicolumn{4}{c|}{GLM-5.2} & \multicolumn{4}{c|}{MistS-4} & \multicolumn{4}{c|}{MistM-3.5} & \multicolumn{4}{c|}{Nem3-S} & \multicolumn{4}{c|}{Nem3-U} \\
 Domain & Pr & Re & $F_1$ & TR & Pr & Re & $F_1$ & TR & Pr & Re & $F_1$ & TR & Pr & Re & $F_1$ & TR & Pr & Re & $F_1$ & TR & Pr & Re & $F_1$ & TR & Pr & Re & $F_1$ & TR \\
\hline
Floortile & \textbf{1} & \textbf{1} & \textbf{1} & \textbf{1} & \textbf{1} & \textbf{1} & \textbf{1} & \textbf{1} & \textbf{1} & \textbf{1} & \textbf{1} & \textbf{1} & .50 & .50 & .50 & 0 & .67 & \textbf{1} & .80 & .83 & .80 & \textbf{1} & .89 & \textbf{1} & .67 & \textbf{1} & .80 & \textbf{1} \\
Freecell & .25 & .50 & .33 & .07 & \textbf{1} & .75 & .86 & .51 & \textbf{1} & \textbf{1} & \textbf{1} & \textbf{1} & 0 & 0 & 0 & 0 & .12 & .25 & .17 & 0 & .04 & .50 & .07 & 0 & .33 & .50 & .40 & .01 \\
GED & .75 & .30 & .43 & 0 & .80 & .40 & .53 & .59 & .83 & \textbf{.50} & \textbf{.62} & \textbf{.69} & 0 & 0 & 0 & .55 & .04 & .10 & .05 & 0 & .22 & .20 & .21 & .31 & .75 & .30 & .43 & \textbf{.69} \\
Hiking & .43 & .75 & .55 & 0 & \textbf{1} & .75 & .86 & \textbf{1} & \textbf{1} & .75 & .86 & \textbf{1} & .33 & .50 & .40 & 0 & \textbf{1} & .75 & .86 & \textbf{1} & .29 & .50 & .36 & 0 & \textbf{1} & \textbf{1} & \textbf{1} & \textbf{1} \\
Logistics00 & .67 & \textbf{1} & .80 & \textbf{1} & \textbf{1} & \textbf{1} & \textbf{1} & \textbf{1} & .80 & \textbf{1} & .89 & \textbf{1} & .25 & .25 & .25 & .03 & 0 & 0 & 0 & 0 & .75 & .75 & .75 & .03 & .67 & \textbf{1} & .80 & \textbf{1} \\
Scanalyzer & .33 & \textbf{1} & .50 & \textbf{1} & \textbf{1} & \textbf{1} & \textbf{1} & \textbf{1} & \textbf{1} & \textbf{1} & \textbf{1} & \textbf{1} & 0 & 0 & 0 & 0 & .17 & .50 & .25 & .93 & .29 & \textbf{1} & .44 & \textbf{1} & .33 & \textbf{1} & .50 & \textbf{1} \\
Slitherlink & .50 & \textbf{1} & .67 & \textbf{1} & \textbf{1} & \textbf{1} & \textbf{1} & \textbf{1} & \textbf{1} & \textbf{1} & \textbf{1} & \textbf{1} & 0 & 0 & 0 & 0 & .17 & .50 & .25 & .33 & \textbf{1} & \textbf{1} & \textbf{1} & \textbf{1} & .50 & \textbf{1} & .67 & \textbf{1} \\
Sokoban & \textbf{1} & \textbf{1} & \textbf{1} & \textbf{1} & \textbf{1} & .50 & .67 & .60 & \textbf{1} & \textbf{1} & \textbf{1} & \textbf{1} & 0 & 0 & 0 & 0 & 0 & 0 & 0 & 0 & .50 & .50 & .50 & .60 & \textbf{1} & \textbf{1} & \textbf{1} & \textbf{1} \\
Tetris & \textbf{1} & .50 & .67 & .50 & \textbf{1} & .50 & .67 & .50 & \textbf{1} & .50 & .67 & .50 & 0 & 0 & 0 & .10 & .27 & \textbf{.75} & .40 & 0 & .11 & .50 & .18 & 0 & .75 & \textbf{.75} & \textbf{.75} & \textbf{1} \\
Thoughtful & .12 & .10 & .11 & 0 & \textbf{.83} & .50 & .62 & 0 & .62 & .50 & .56 & \textbf{.06} & 0 & 0 & 0 & 0 & .26 & \textbf{.60} & .36 & \textbf{.06} & .06 & .10 & .07 & 0 & .67 & \textbf{.60} & \textbf{.63} & \textbf{.06} \\
Tidybot & .44 & \textbf{.67} & .53 & \textbf{1} & \textbf{.73} & \textbf{.67} & \textbf{.70} & \textbf{1} & .70 & .58 & .64 & \textbf{1} & .50 & .17 & .25 & .44 & .38 & .25 & .30 & \textbf{1} & .40 & .17 & .24 & 0 & .44 & .33 & .38 & \textbf{1} \\
Woodwork11 & .40 & \textbf{1} & .57 & .12 & .50 & \textbf{1} & .67 & .12 & .86 & \textbf{1} & \textbf{.92} & \textbf{1} & .15 & .33 & .21 & .62 & \textbf{1} & .67 & .80 & .88 & .36 & .83 & .50 & 0 & .40 & .67 & .50 & .12 \\
\hline
AVG & .57 & .73 & .60 & .56 & \textbf{.91} & .76 & .80 & .69 & .90 & \textbf{.82} & \textbf{.85} & \textbf{.85} & .14 & .15 & .13 & .15 & .34 & .45 & .35 & .42 & .40 & .59 & .43 & .33 & .63 & .76 & .65 & .74 \\
STD & .29 & .31 & .25 & .46 & .15 & .23 & .17 & .35 & .13 & .22 & .17 & .28 & .19 & .19 & .17 & .23 & .34 & .31 & .30 & .44 & .29 & .32 & .30 & .42 & .22 & .26 & .21 & .40 \\
\hline
\end{tabular}
}%
\ifdim\wd0>\textwidth
\resizebox{\textwidth}{!}{\usebox0}%
\else
\usebox0
\fi
\endgroup
\caption{
    Per-domain results using the \notrace{} prompt in Figure~\ref{fig:prompt-variations}; test traces are not sent to the LLM. The default effort setting is used for each LLM.
    TR is the test pass rate, i.e., the fraction of tests satisfied by the repair; the AVG row averages this rate across domains.
    For each domain and the AVG row, the best precision, recall, $F_1$, and TR values across all models are shown in bold.
    The upper and lower parts of the table contain models released in 2025 and 2026, respectively.
}
\label{tab:new-all-twin-models-notrace}
\end{table*}

%% file: Tabs/trace.tex
\begin{table*}[p]
\centering
\setlength{\tabcolsep}{3pt}
\renewcommand{\arraystretch}{1.2}
\begingroup
\setbox0=\hbox{%
\begin{tabular}{|l|cccc|cccc|cccc|cccc|cccc|cccc|cccc|}
\hline
 & \multicolumn{4}{c|}{OSS20B} & \multicolumn{4}{c|}{OSS120B} & \multicolumn{4}{c|}{Minst-14B} & \multicolumn{4}{c|}{MistL-3} & \multicolumn{4}{c|}{Phi4} & \multicolumn{4}{c|}{Qwen3-30B} & \multicolumn{4}{c|}{Qwen3-32B} \\
 Domain & Pr & Re & $F_1$ & TR & Pr & Re & $F_1$ & TR & Pr & Re & $F_1$ & TR & Pr & Re & $F_1$ & TR & Pr & Re & $F_1$ & TR & Pr & Re & $F_1$ & TR & Pr & Re & $F_1$ & TR \\
\hline
Floortile & .67 & .50 & .57 & \textbf{1} & \textbf{1} & .75 & .86 & \textbf{1} & .25 & .25 & .25 & .83 & \textbf{1} & .75 & .86 & \textbf{1} & 0 & 0 & 0 & .83 & .67 & .50 & .57 & 0 & .50 & .50 & .50 & .83 \\
Freecell & $e_1$ & $e_1$ & $e_1$ & 0 & $e_1$ & $e_1$ & $e_1$ & 0 & 0 & 0 & 0 & .01 & .03 & \textbf{.75} & .05 & .01 & $e_1$ & $e_1$ & $e_1$ & 0 & $e_1$ & $e_1$ & $e_1$ & 0 & $e_1$ & $e_1$ & $e_1$ & 0 \\
GED & .07 & .10 & .08 & .31 & .50 & .10 & .17 & .31 & 0 & 0 & 0 & 0 & \textbf{1} & .10 & .18 & 0 & 0 & 0 & 0 & 0 & 0 & 0 & 0 & 0 & 0 & 0 & 0 & 0 \\
Hiking & .67 & .50 & .57 & \textbf{1} & \textbf{1} & \textbf{.75} & \textbf{.86} & \textbf{1} & .17 & .25 & .20 & 0 & .50 & .50 & .50 & .22 & 0 & 0 & 0 & 0 & \textbf{1} & .50 & .67 & .22 & .67 & .50 & .57 & .22 \\
Logistics00 & .50 & .50 & .50 & .03 & \textbf{1} & \textbf{1} & \textbf{1} & \textbf{1} & .14 & .25 & .18 & .03 & .75 & .75 & .75 & .03 & $e_1$ & $e_1$ & $e_1$ & 0 & .33 & .25 & .29 & 0 & 0 & 0 & 0 & 0 \\
Scanalyzer & .14 & .50 & .22 & .07 & \textbf{1} & .50 & .67 & .93 & .09 & .50 & .15 & .93 & .20 & .50 & .29 & .93 & $e_1$ & $e_1$ & $e_1$ & 0 & \textbf{1} & .50 & .67 & .93 & 0 & 0 & 0 & 0 \\
Slitherlink & \textbf{1} & .50 & .67 & .67 & \textbf{1} & \textbf{1} & \textbf{1} & \textbf{1} & 0 & 0 & 0 & 0 & .50 & .50 & .50 & .33 & 0 & 0 & 0 & 0 & 0 & 0 & 0 & 0 & .33 & .50 & .40 & .33 \\
Sokoban & 0 & 0 & 0 & 0 & .33 & .50 & .40 & .40 & 0 & 0 & 0 & 0 & 0 & 0 & 0 & 0 & $e_1$ & $e_1$ & $e_1$ & 0 & 0 & 0 & 0 & 0 & \textbf{1} & .50 & .67 & .40 \\
Tetris & .08 & .25 & .12 & 0 & .50 & .25 & .33 & \textbf{.50} & 0 & 0 & 0 & 0 & .11 & .25 & .15 & 0 & $e_1$ & $e_1$ & $e_1$ & 0 & 0 & 0 & 0 & 0 & .25 & \textbf{.50} & .33 & \textbf{.50} \\
Thoughtful & 0 & 0 & 0 & 0 & .02 & .10 & .04 & \textbf{.06} & .08 & .10 & .09 & \textbf{.06} & .05 & .20 & .08 & 0 & $e_1$ & $e_1$ & $e_1$ & 0 & $e_3$ & $e_3$ & $e_3$ & 0 & 0 & 0 & 0 & 0 \\
Tidybot & 0 & 0 & 0 & 0 & \textbf{1} & .08 & .15 & .56 & 0 & 0 & 0 & 0 & .83 & .42 & .56 & \textbf{1} & $e_1$ & $e_1$ & $e_1$ & 0 & 0 & 0 & 0 & 0 & .33 & .17 & .22 & .44 \\
Woodwork11 & 0 & 0 & 0 & 0 & .67 & .33 & .44 & .12 & .33 & .33 & .33 & .12 & .25 & .33 & .29 & .12 & $e_1$ & $e_1$ & $e_1$ & 0 & 0 & 0 & 0 & 0 & 0 & 0 & 0 & 0 \\
\hline
AVG & .26 & .24 & .23 & .26 & .67 & .45 & .49 & .57 & .09 & .14 & .10 & .17 & .44 & .42 & .35 & .30 & 0 & 0 & 0 & .07 & .25 & .15 & .18 & .10 & .26 & .22 & .22 & .23 \\
STD & .34 & .23 & .26 & .38 & .38 & .34 & .36 & .38 & .11 & .16 & .11 & .32 & .37 & .24 & .27 & .40 & 0 & 0 & 0 & .23 & .39 & .22 & .27 & .26 & .31 & .24 & .25 & .26 \\
\hline
\end{tabular}
}%
\ifdim\wd0>\textwidth
\resizebox{\textwidth}{!}{\usebox0}%
\else
\usebox0
\fi
\endgroup

\par\vspace{6pt}
\begingroup
\setbox0=\hbox{%
\begin{tabular}{|l|cccc|cccc|cccc|cccc|cccc|cccc|cccc|}
\hline
 & \multicolumn{4}{c|}{DSV4-F} & \multicolumn{4}{c|}{DSV4-P} & \multicolumn{4}{c|}{GLM-5.2} & \multicolumn{4}{c|}{MistS-4} & \multicolumn{4}{c|}{MistM-3.5} & \multicolumn{4}{c|}{Nem3-S} & \multicolumn{4}{c|}{Nem3-U} \\
 Domain & Pr & Re & $F_1$ & TR & Pr & Re & $F_1$ & TR & Pr & Re & $F_1$ & TR & Pr & Re & $F_1$ & TR & Pr & Re & $F_1$ & TR & Pr & Re & $F_1$ & TR & Pr & Re & $F_1$ & TR \\
\hline
Floortile & \textbf{1} & \textbf{1} & \textbf{1} & \textbf{1} & \textbf{1} & \textbf{1} & \textbf{1} & \textbf{1} & \textbf{1} & \textbf{1} & \textbf{1} & \textbf{1} & .67 & .50 & .57 & .83 & .50 & .75 & .60 & .83 & \textbf{1} & .25 & .40 & .83 & \textbf{1} & \textbf{1} & \textbf{1} & \textbf{1} \\
Freecell & .33 & .25 & .29 & .21 & \textbf{1} & \textbf{.75} & \textbf{.86} & \textbf{.70} & \textbf{1} & \textbf{.75} & \textbf{.86} & \textbf{.70} & .02 & .25 & .04 & .51 & .33 & .50 & .40 & .21 & 0 & 0 & 0 & 0 & \textbf{1} & .50 & .67 & .21 \\
GED & .78 & \textbf{.70} & \textbf{.74} & \textbf{1} & .75 & .30 & .43 & \textbf{1} & .67 & .20 & .31 & .97 & .12 & .10 & .11 & 0 & \textbf{1} & .10 & .18 & .31 & .50 & .10 & .17 & .31 & .75 & .30 & .43 & \textbf{1} \\
Hiking & \textbf{1} & \textbf{.75} & \textbf{.86} & \textbf{1} & \textbf{1} & \textbf{.75} & \textbf{.86} & \textbf{1} & \textbf{1} & \textbf{.75} & \textbf{.86} & \textbf{1} & 0 & 0 & 0 & 0 & \textbf{1} & \textbf{.75} & \textbf{.86} & \textbf{1} & .67 & .50 & .57 & .78 & \textbf{1} & \textbf{.75} & \textbf{.86} & \textbf{1} \\
Logistics00 & .67 & \textbf{1} & .80 & \textbf{1} & .67 & \textbf{1} & .80 & \textbf{1} & \textbf{1} & \textbf{1} & \textbf{1} & \textbf{1} & 0 & 0 & 0 & 0 & \textbf{1} & .50 & .67 & 0 & .60 & .75 & .67 & \textbf{1} & \textbf{1} & \textbf{1} & \textbf{1} & \textbf{1} \\
Scanalyzer & \textbf{1} & \textbf{1} & \textbf{1} & \textbf{1} & \textbf{1} & \textbf{1} & \textbf{1} & \textbf{1} & \textbf{1} & \textbf{1} & \textbf{1} & \textbf{1} & 0 & 0 & 0 & 0 & .20 & .50 & .29 & .93 & \textbf{1} & .50 & .67 & .93 & \textbf{1} & \textbf{1} & \textbf{1} & \textbf{1} \\
Slitherlink & .50 & \textbf{1} & .67 & \textbf{1} & \textbf{1} & \textbf{1} & \textbf{1} & \textbf{1} & \textbf{1} & \textbf{1} & \textbf{1} & \textbf{1} & 0 & 0 & 0 & 0 & .20 & .50 & .29 & .67 & \textbf{1} & .50 & .67 & .67 & \textbf{1} & \textbf{1} & \textbf{1} & \textbf{1} \\
Sokoban & \textbf{1} & \textbf{1} & \textbf{1} & \textbf{1} & \textbf{1} & .50 & .67 & .60 & \textbf{1} & .50 & .67 & .60 & .33 & .50 & .40 & 0 & .20 & .50 & .29 & .40 & 0 & 0 & 0 & 0 & \textbf{1} & \textbf{1} & \textbf{1} & \textbf{1} \\
Tetris & 0 & 0 & 0 & 0 & \textbf{1} & \textbf{.50} & \textbf{.67} & \textbf{.50} & \textbf{1} & \textbf{.50} & \textbf{.67} & \textbf{.50} & 0 & 0 & 0 & 0 & .06 & .25 & .09 & \textbf{.50} & $e_3$ & $e_3$ & $e_3$ & 0 & \textbf{1} & .25 & .40 & \textbf{.50} \\
Thoughtful & \textbf{1} & .20 & .33 & 0 & .67 & \textbf{.60} & .63 & \textbf{.06} & .83 & .50 & .62 & \textbf{.06} & 0 & 0 & 0 & 0 & 0 & 0 & 0 & 0 & .11 & .10 & .11 & 0 & \textbf{1} & .50 & \textbf{.67} & 0 \\
Tidybot & .50 & .25 & .33 & \textbf{1} & .86 & \textbf{.50} & \textbf{.63} & \textbf{1} & .80 & .33 & .47 & \textbf{1} & 0 & 0 & 0 & 0 & .11 & .08 & .10 & .44 & \textbf{1} & .08 & .15 & .56 & .60 & .25 & .35 & \textbf{1} \\
Woodwork11 & .50 & .50 & .50 & .75 & .27 & .67 & .38 & \textbf{1} & \textbf{1} & \textbf{.83} & \textbf{.91} & \textbf{1} & 0 & 0 & 0 & 0 & .50 & \textbf{.83} & .62 & .25 & \textbf{1} & .50 & .67 & .62 & \textbf{1} & .33 & .50 & .62 \\
\hline
AVG & .69 & .64 & .63 & .75 & .85 & \textbf{.71} & .74 & \textbf{.82} & .94 & .70 & \textbf{.78} & .82 & .10 & .11 & .09 & .11 & .42 & .44 & .36 & .46 & .57 & .27 & .34 & .48 & \textbf{.95} & .66 & .74 & .78 \\
STD & .32 & .36 & .32 & .40 & .22 & .23 & .20 & .29 & .11 & .27 & .22 & .29 & .20 & .19 & .18 & .26 & .36 & .26 & .26 & .32 & .42 & .25 & .28 & .38 & .12 & .32 & .26 & .34 \\
\hline
\end{tabular}
}%
\ifdim\wd0>\textwidth
\resizebox{\textwidth}{!}{\usebox0}%
\else
\usebox0
\fi
\endgroup
\caption{
    Per-domain results using the \withtrace{} prompt in Figure~\ref{fig:prompt-variations}; test traces are sent to the LLM.
    The default effort setting is used for each LLM.
    TR is the test pass rate, i.e., the fraction of tests satisfied by the repair; the AVG row averages this rate across domains.
    For each domain and the AVG row, the best precision, recall, $F_1$, and TR values across all models are shown in bold.
    The upper part of the table contains models released in 2025, and the lower part contains models released in 2026.
    Missing values are marked as \(e_1\) for context-limit errors, \(e_2\) for other API-call errors, and \(e_3\) for invalid or unparsable output.
    Missing data are treated as 0 in our calculations.
}
\label{tab:new-all-twin-models-trace}
\end{table*}

%% file: Tabs/notrace_reasoning.tex
\begin{table*}[p]
\centering
\setlength{\tabcolsep}{3pt}
\renewcommand{\arraystretch}{1.2}
\begingroup
\setbox0=\hbox{%
\begin{tabular}{|l|cccc|cccc|cccc|}
\hline
 & \multicolumn{4}{c|}{OSS20B} & \multicolumn{4}{c|}{OSS120B} & \multicolumn{4}{c|}{Qwen3-32B} \\
 Domain & Pr & Re & $F_1$ & TR & Pr & Re & $F_1$ & TR & Pr & Re & $F_1$ & TR \\
\hline
Floortile & .67 & \textbf{1} & .80 & .83 & .67 & \textbf{1} & .80 & .83 & .60 & .75 & .67 & .83 \\
Freecell & .14 & .50 & .22 & .21 & $e_3$ & $e_3$ & $e_3$ & 0 & .12 & .25 & .17 & 0 \\
GED & .06 & .10 & .07 & 0 & $e_3$ & $e_3$ & $e_3$ & 0 & .25 & .10 & .14 & 0 \\
Hiking & \textbf{1} & \textbf{.75} & \textbf{.86} & \textbf{1} & \textbf{1} & \textbf{.75} & \textbf{.86} & \textbf{1} & .67 & .50 & .57 & \textbf{1} \\
Logistics00 & \textbf{1} & \textbf{1} & \textbf{1} & \textbf{1} & .80 & \textbf{1} & .89 & \textbf{1} & \textbf{1} & .75 & .86 & .97 \\
Scanalyzer & .33 & \textbf{1} & .50 & \textbf{1} & .11 & .50 & .18 & 0 & .20 & \textbf{1} & .33 & \textbf{1} \\
Slitherlink & \textbf{1} & \textbf{1} & \textbf{1} & \textbf{1} & \textbf{1} & \textbf{1} & \textbf{1} & \textbf{1} & .33 & \textbf{1} & .50 & \textbf{1} \\
Sokoban & \textbf{1} & \textbf{1} & \textbf{1} & \textbf{1} & .50 & \textbf{1} & .67 & \textbf{1} & .33 & .50 & .40 & .40 \\
Tetris & .25 & .25 & .25 & 0 & $e_3$ & $e_3$ & $e_3$ & 0 & .20 & .50 & .29 & .50 \\
Thoughtful & .02 & .10 & .03 & 0 & $e_3$ & $e_3$ & $e_3$ & 0 & 0 & 0 & 0 & 0 \\
Tidybot & 0 & 0 & 0 & 0 & $e_3$ & $e_3$ & $e_3$ & 0 & .20 & .08 & .12 & .44 \\
Woodwork11 & .20 & .50 & .29 & 0 & $e_3$ & $e_3$ & $e_3$ & 0 & \textbf{1} & .67 & \textbf{.80} & \textbf{.88} \\
\hline
AVG & .47 & .60 & .50 & .50 & .34 & .44 & .37 & .40 & .41 & .51 & .40 & .58 \\
STD & .41 & .39 & .39 & .47 & .40 & .46 & .41 & .48 & .32 & .33 & .27 & .40 \\
\hline
\end{tabular}
}%
\ifdim\wd0>\textwidth
\resizebox{\textwidth}{!}{\usebox0}%
\else
\usebox0
\fi
\endgroup

\par\vspace{6pt}
\begingroup
\setbox0=\hbox{%
\begin{tabular}{|l|cccc|cccc|cccc|cccc|cccc|}
\hline
 & \multicolumn{4}{c|}{DSV4-F} & \multicolumn{4}{c|}{DSV4-P} & \multicolumn{4}{c|}{GLM-5.2} & \multicolumn{4}{c|}{MistS-4} & \multicolumn{4}{c|}{MistM-3.5} \\
 Domain & Pr & Re & $F_1$ & TR & Pr & Re & $F_1$ & TR & Pr & Re & $F_1$ & TR & Pr & Re & $F_1$ & TR & Pr & Re & $F_1$ & TR \\
\hline
Floortile & \textbf{1} & \textbf{1} & \textbf{1} & \textbf{1} & .80 & \textbf{1} & .89 & \textbf{1} & \textbf{1} & \textbf{1} & \textbf{1} & \textbf{1} & .67 & .50 & .57 & 0 & .67 & \textbf{1} & .80 & .83 \\
Freecell & \textbf{1} & .50 & .67 & .21 & \textbf{1} & \textbf{1} & \textbf{1} & \textbf{1} & \textbf{1} & \textbf{1} & \textbf{1} & \textbf{1} & .25 & .50 & .33 & .21 & .18 & .50 & .27 & 0 \\
GED & .78 & \textbf{.70} & .74 & \textbf{.69} & .88 & \textbf{.70} & \textbf{.78} & .59 & \textbf{1} & .50 & .67 & .55 & .50 & .20 & .29 & 0 & .62 & .50 & .56 & .10 \\
Hiking & .75 & \textbf{.75} & .75 & \textbf{1} & .33 & \textbf{.75} & .46 & 0 & \textbf{1} & \textbf{.75} & \textbf{.86} & \textbf{1} & .17 & \textbf{.75} & .27 & 0 & .43 & \textbf{.75} & .55 & 0 \\
Logistics00 & .67 & \textbf{1} & .80 & \textbf{1} & .50 & \textbf{1} & .67 & \textbf{1} & .67 & \textbf{1} & .80 & \textbf{1} & .60 & .75 & .67 & \textbf{1} & .50 & .75 & .60 & .03 \\
Scanalyzer & .33 & \textbf{1} & .50 & \textbf{1} & \textbf{1} & \textbf{1} & \textbf{1} & \textbf{1} & \textbf{1} & \textbf{1} & \textbf{1} & \textbf{1} & .33 & \textbf{1} & .50 & \textbf{1} & .33 & \textbf{1} & .50 & \textbf{1} \\
Slitherlink & \textbf{1} & \textbf{1} & \textbf{1} & \textbf{1} & .10 & \textbf{1} & .18 & 0 & \textbf{1} & \textbf{1} & \textbf{1} & \textbf{1} & .33 & .50 & .40 & .33 & .50 & \textbf{1} & .67 & \textbf{1} \\
Sokoban & .50 & \textbf{1} & .67 & \textbf{1} & \textbf{1} & \textbf{1} & \textbf{1} & \textbf{1} & \textbf{1} & \textbf{1} & \textbf{1} & \textbf{1} & .33 & .50 & .40 & .60 & \textbf{1} & \textbf{1} & \textbf{1} & \textbf{1} \\
Tetris & .75 & \textbf{.75} & .75 & .50 & .50 & \textbf{.75} & .60 & \textbf{1} & \textbf{1} & \textbf{.75} & \textbf{.86} & .50 & .20 & .50 & .29 & 0 & .11 & .50 & .17 & \textbf{1} \\
Thoughtful & .28 & \textbf{.70} & .40 & \textbf{.06} & .33 & \textbf{.70} & .45 & \textbf{.06} & \textbf{1} & .60 & \textbf{.75} & \textbf{.06} & .25 & .30 & .27 & \textbf{.06} & .22 & .40 & .29 & \textbf{.06} \\
Tidybot & .67 & .67 & .67 & \textbf{1} & .53 & \textbf{.75} & .62 & \textbf{1} & \textbf{.75} & \textbf{.75} & \textbf{.75} & \textbf{1} & \textbf{.75} & .25 & .38 & .44 & .17 & .42 & .24 & .44 \\
Woodwork11 & .29 & .83 & .43 & .12 & .40 & \textbf{1} & .57 & .12 & .80 & .67 & .73 & .75 & \textbf{1} & .33 & .50 & .62 & .50 & .83 & .62 & .75 \\
\hline
AVG & .67 & .83 & .70 & .72 & .61 & \textbf{.89} & .69 & .65 & \textbf{.93} & .83 & \textbf{.87} & \textbf{.82} & .45 & .51 & .41 & .36 & .44 & .72 & .52 & .52 \\
STD & .26 & .16 & .18 & .37 & .30 & .13 & .25 & .44 & .12 & .18 & .12 & .29 & .25 & .22 & .12 & .36 & .25 & .24 & .24 & .43 \\
\hline
\end{tabular}
}%
\ifdim\wd0>\textwidth
\resizebox{\textwidth}{!}{\usebox0}%
\else
\usebox0
\fi
\endgroup
\caption{
    Per-domain results using the \notrace{} prompt in Figure~\ref{fig:prompt-variations}; test traces are not sent to the LLM.
    The high-effort setting is used for each LLM when available (see Table~\ref{tab:new-model-based}); models without this option are omitted.
    TR is the test pass rate, i.e., the fraction of tests satisfied by the repair; the AVG row averages this rate across domains.
    For each domain and the AVG row, the best precision, recall, $F_1$, and TR values across all models are shown in bold.
    The upper part of the table contains models released in 2025, and the lower part contains models released in 2026.
    Missing values are marked as \(e_1\) for context-limit errors, \(e_2\) for other API-call errors, and \(e_3\) for invalid or unparsable output.
    Missing data are treated as 0 in our calculations.
}
\label{tab:new-all-twin-models-notrace-high}
\end{table*}

%% file: Tabs/trace_reasoning.tex
\begin{table*}[p]
\centering
\setlength{\tabcolsep}{3pt}
\renewcommand{\arraystretch}{1.2}
\begingroup
\setbox0=\hbox{%
\begin{tabular}{|l|cccc|cccc|cccc|}
\hline
 & \multicolumn{4}{c|}{OSS20B} & \multicolumn{4}{c|}{OSS120B} & \multicolumn{4}{c|}{Qwen3-32B} \\
 Domain & Pr & Re & $F_1$ & TR & Pr & Re & $F_1$ & TR & Pr & Re & $F_1$ & TR \\
\hline
Floortile & \textbf{1} & \textbf{1} & \textbf{1} & \textbf{1} & $e_3$ & $e_3$ & $e_3$ & 0 & .25 & .25 & .25 & 0 \\
Freecell & $e_1$ & $e_1$ & $e_1$ & 0 & $e_1$ & $e_1$ & $e_1$ & 0 & $e_1$ & $e_1$ & $e_1$ & 0 \\
GED & .06 & .10 & .08 & .55 & $e_3$ & $e_3$ & $e_3$ & 0 & 0 & 0 & 0 & 0 \\
Hiking & .60 & \textbf{.75} & .67 & \textbf{1} & $e_3$ & $e_3$ & $e_3$ & 0 & .67 & .50 & .57 & \textbf{1} \\
Logistics00 & .75 & .75 & .75 & \textbf{1} & .50 & .75 & .60 & .03 & 0 & 0 & 0 & .03 \\
Scanalyzer & .33 & \textbf{1} & .50 & \textbf{1} & \textbf{1} & \textbf{1} & \textbf{1} & \textbf{1} & 0 & 0 & 0 & 0 \\
Slitherlink & \textbf{1} & \textbf{1} & \textbf{1} & \textbf{1} & $e_3$ & $e_3$ & $e_3$ & 0 & .50 & .50 & .50 & .33 \\
Sokoban & .33 & .50 & .40 & .60 & $e_3$ & $e_3$ & $e_3$ & 0 & 0 & 0 & 0 & 0 \\
Tetris & .50 & .50 & .50 & .50 & $e_3$ & $e_3$ & $e_3$ & 0 & 0 & 0 & 0 & 0 \\
Thoughtful & 0 & 0 & 0 & 0 & .33 & .20 & .25 & \textbf{.06} & $e_2$ & $e_2$ & $e_2$ & 0 \\
Tidybot & 0 & 0 & 0 & 0 & $e_3$ & $e_3$ & $e_3$ & 0 & 0 & 0 & 0 & 0 \\
Woodwork11 & 0 & 0 & 0 & 0 & $e_3$ & $e_3$ & $e_3$ & 0 & 0 & 0 & 0 & 0 \\
\hline
AVG & .38 & .47 & .41 & .55 & .15 & .16 & .15 & .09 & .12 & .10 & .11 & .11 \\
STD & .37 & .41 & .37 & .43 & .30 & .33 & .31 & .27 & .22 & .19 & .20 & .28 \\
\hline
\end{tabular}
}%
\ifdim\wd0>\textwidth
\resizebox{\textwidth}{!}{\usebox0}%
\else
\usebox0
\fi
\endgroup

\par\vspace{6pt}
\begingroup
\setbox0=\hbox{%
\begin{tabular}{|l|cccc|cccc|cccc|cccc|cccc|}
\hline
 & \multicolumn{4}{c|}{DSV4-F} & \multicolumn{4}{c|}{DSV4-P} & \multicolumn{4}{c|}{GLM-5.2} & \multicolumn{4}{c|}{MistS-4} & \multicolumn{4}{c|}{MistM-3.5} \\
 Domain & Pr & Re & $F_1$ & TR & Pr & Re & $F_1$ & TR & Pr & Re & $F_1$ & TR & Pr & Re & $F_1$ & TR & Pr & Re & $F_1$ & TR \\
\hline
Floortile & \textbf{1} & \textbf{1} & \textbf{1} & \textbf{1} & .67 & \textbf{1} & .80 & \textbf{1} & \textbf{1} & \textbf{1} & \textbf{1} & \textbf{1} & \textbf{1} & .75 & .86 & .83 & .67 & \textbf{1} & .80 & \textbf{1} \\
Freecell & \textbf{1} & .50 & .67 & .21 & .80 & \textbf{1} & .89 & \textbf{1} & \textbf{1} & \textbf{1} & \textbf{1} & \textbf{1} & .02 & .50 & .03 & .01 & $e_3$ & $e_3$ & $e_3$ & 0 \\
GED & \textbf{.83} & \textbf{.50} & \textbf{.62} & \textbf{1} & .80 & .40 & .53 & \textbf{1} & .80 & .40 & .53 & \textbf{1} & 0 & 0 & 0 & 0 & .75 & .30 & .43 & \textbf{1} \\
Hiking & \textbf{1} & \textbf{.75} & \textbf{.86} & \textbf{1} & \textbf{1} & \textbf{.75} & \textbf{.86} & \textbf{1} & \textbf{1} & \textbf{.75} & \textbf{.86} & \textbf{1} & .75 & \textbf{.75} & .75 & \textbf{1} & .75 & \textbf{.75} & .75 & \textbf{1} \\
Logistics00 & .80 & \textbf{1} & .89 & \textbf{1} & .67 & \textbf{1} & .80 & \textbf{1} & .80 & \textbf{1} & .89 & \textbf{1} & .75 & .75 & .75 & \textbf{1} & \textbf{1} & \textbf{1} & \textbf{1} & \textbf{1} \\
Scanalyzer & \textbf{1} & \textbf{1} & \textbf{1} & \textbf{1} & \textbf{1} & \textbf{1} & \textbf{1} & \textbf{1} & \textbf{1} & \textbf{1} & \textbf{1} & \textbf{1} & \textbf{1} & \textbf{1} & \textbf{1} & \textbf{1} & \textbf{1} & \textbf{1} & \textbf{1} & \textbf{1} \\
Slitherlink & \textbf{1} & \textbf{1} & \textbf{1} & \textbf{1} & \textbf{1} & \textbf{1} & \textbf{1} & \textbf{1} & \textbf{1} & .50 & .67 & .67 & .50 & \textbf{1} & .67 & \textbf{1} & \textbf{1} & .50 & .67 & .67 \\
Sokoban & \textbf{1} & \textbf{1} & \textbf{1} & \textbf{1} & \textbf{1} & \textbf{1} & \textbf{1} & \textbf{1} & \textbf{1} & \textbf{1} & \textbf{1} & \textbf{1} & \textbf{1} & .50 & .67 & .60 & \textbf{1} & \textbf{1} & \textbf{1} & \textbf{1} \\
Tetris & \textbf{1} & .25 & .40 & .50 & \textbf{1} & \textbf{.75} & \textbf{.86} & \textbf{1} & \textbf{1} & .50 & .67 & .50 & 0 & 0 & 0 & 0 & \textbf{1} & .50 & .67 & .50 \\
Thoughtful & .27 & .30 & .29 & \textbf{.06} & .55 & .60 & .57 & \textbf{.06} & .89 & \textbf{.80} & \textbf{.84} & \textbf{.06} & \textbf{1} & .20 & .33 & 0 & .09 & .10 & .10 & 0 \\
Tidybot & .57 & .33 & .42 & \textbf{1} & .67 & \textbf{.50} & \textbf{.57} & \textbf{1} & \textbf{.80} & .33 & .47 & \textbf{1} & .50 & .08 & .14 & .56 & .67 & .17 & .27 & \textbf{1} \\
Woodwork11 & .27 & .67 & .38 & \textbf{1} & \textbf{1} & \textbf{1} & \textbf{1} & \textbf{1} & \textbf{1} & .83 & .91 & \textbf{1} & .08 & .33 & .13 & 0 & \textbf{1} & .67 & .80 & .88 \\
\hline
AVG & .81 & .69 & .71 & .81 & .85 & \textbf{.83} & \textbf{.82} & \textbf{.92} & \textbf{.94} & .76 & .82 & .85 & .55 & .49 & .44 & .50 & .74 & .58 & .62 & .75 \\
STD & .27 & .29 & .27 & .33 & .17 & .22 & .17 & .26 & .09 & .25 & .18 & .29 & .41 & .35 & .36 & .44 & .34 & .36 & .33 & .37 \\
\hline
\end{tabular}
}%
\ifdim\wd0>\textwidth
\resizebox{\textwidth}{!}{\usebox0}%
\else
\usebox0
\fi
\endgroup
\caption{
    Per-domain results using the \withtrace{} prompt in Figure~\ref{fig:prompt-variations}; test traces are sent to the LLM.
    The high-effort setting is used for each LLM when available (see Table~\ref{tab:new-model-based}); models without this option are omitted.
    TR is the test pass rate, i.e., the fraction of tests satisfied by the repair; the AVG row averages this rate across domains.
    For each domain and the AVG row, the best precision, recall, $F_1$, and TR values across all models are shown in bold.
    The upper part of the table contains models released in 2025, and the lower part contains models released in 2026.
    Missing values are marked as \(e_1\) for context-limit errors, \(e_2\) for other API-call errors, and \(e_3\) for invalid or unparsable output.
    Missing data are treated as 0 in our calculations.
}
\label{tab:new-all-twin-models-trace-high}
\end{table*}